\documentclass{article}

\usepackage{graphicx} 
\usepackage{subfigure}

\usepackage{natbib}

\usepackage{algorithm}
\usepackage{algorithmic}

\usepackage{hyperref}

\usepackage[accepted]{icml2013}

\icmltitlerunning{A Novel Hybrid Machine Learning Model for Auto-Classification of Retinal Diseases}

\begin{document} 

\twocolumn[
\icmltitle{A Novel Hybrid Machine Learning Model for Auto-Classification of Retinal Diseases\\ }

\icmlauthor{C.-H. Huck Yang* ; Jia-Hong Huang* ; Fangyu Liu ; Fang-Yi Chiu ; Mengya Gao ; Weifeng Lyu; I-Hung Lin M.D.; Jesper Tegner\\}{chao-han.yang@kaust.edu.sa* ; r03543018@ntu.edu.tw* ; fangyu.liu@uwaterloo.ca ; fachiu@eng.ucsd.edu ; daisygao@gatech.edu ; wlyu6@gatech.edu; petercard@gmail.com ; jesper.tegner@kaust.edu.sa}
\icmladdress{Living Systems Laboratory, BESE, CEMSE, King Abdullah University of Science and Technology, Thuwal, KSA* ; National Taiwan University, Taipei City, Taiwan*; University of Waterloo, ON, Canada ; University of California San Diego, CA, USA.; Georgia Institute of Technology, GA, USA.;  Department of Ophthalmology, Tri-Service General Hospital, National Defense Medical Center, Taiwan}

\icmlkeywords{boring formatting information, machine learning, ICML}

\vskip 0.3in
]

\begin{abstract} 
Automatic clinical diagnosis of retinal diseases has emerged as a promising approach to facilitate discovery in areas with limited access to specialists. We propose a novel visual-assisted diagnosis hybrid model based on the support vector machine (SVM) and deep neural networks (DNNs). The model incorporates complementary strengths of DNNs and SVM. Furthermore, we present a new clinical retina label collection for ophthalmology incorporating 32 retina diseases classes. Using EyeNet, our model achieves 89.73\% diagnosis accuracy and the model performance is comparable to the professional ophthalmologists.
\vspace{-0.3cm}
\end{abstract}

\section{Introduction}


Computational Retinal disease methods \cite{tan2009detection,lalezary2006baseline} has been investigated extensively through different signal processing techniques. Retinal diseases is accessible to machine driven techniques due to their visual nature in contrast other common human diseases requiring invasive techniques for diagnosis or treatments. Typically, the diagnosis accuracy of retinal diseases based on the clinical retinal images is highly dependent on the practical experience of physician or ophthalmologist. However, not every doctor has sufficient practical experience. Therefore, developing an automatic retinal diseases detection system is important and it will broadly facilitate diagnostic accuracy of retinal diseases. For the remote rural area, where there are no ophthalmologists locally to screen retinal disease, the automatic retinal diseases detection system also can help non-ophthalmologists to find the patient of the retinal disease, and further, refer them to the medical center for further treatment.

\begin{figure}
\begin{center}
   \includegraphics[width=1.0\linewidth]{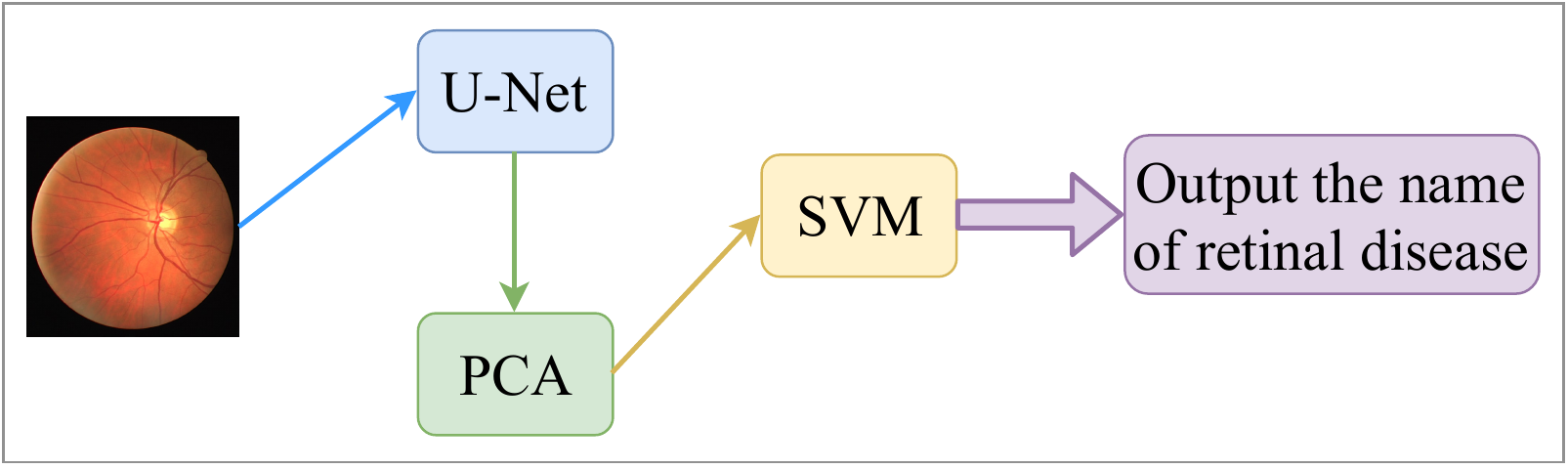}
\end{center}
\vspace{-0.6cm}
   \caption{This figure represents our proposed hybrid-model. A raw retinal image as provided as input of DNNs, U-Net, and then we pass the output of U-Net to the dimension reduction module, PCA. Finally, the output of PCA module sent as input to the retina disease classifier, SVM, which outputs the name of predicted retina disease.}
\label{fig:figure1}
\vspace{-0.4cm}
\end{figure}

The developing of automatic diseases detection (ADD) \cite{sharifi2002classified} alleviate enormous pressure from social healthcare systems. Retinal symptom analysis \cite{abramoff2010retinal} is one of the important ADD applications. Moreover, the increasing number of cases of diabetic retinopathy globally requires extending efforts in developing visual tools to assist in the analytic of the series of retinal disease. These decision support systems for retinal ADD, as \cite{bhattacharya2014watermarking} for non-proliferative diabetic retinopathy have been improved from recent machine learning success on the high dimensional images processing by featuring details on the blood vessel.  \cite{lin2000rotation} demonstrated an automated technique for the segmentation of the blood vessels by tracking the center of the vessels on Kalman Filter. However, these pattern recognition based classification still rely on hand-crafted features and only specify for evaluating single retinal symptom. Despite extensive efforts using wavelet signal processing, retinal ADD remains a viable target for improved machine learning techniques applicable for point-of-care (POC) medical diagnosis and treatment in the aging society \cite{cochocki1993neural}. 

To the best of our knowledge, the amount of clinical retinal images are less compared to other cell imaging data, such as blood cell and a cancer cell. Yet, a vanilla deep learning based diseases diagnosis system requires large amounts of data. Here we, therefore, propose a novel visual-assisted diagnosis algorithm which is based on an integration of support vector machine and deep neural networks. The primary goal of this work is to automatically classify 32 specific retinal diseases for human beings with the reliable clinical-assisted ability on the intelligent medicine approaches. To foster the long-term visual analytics research, we also present a visual clinical label collection, EyeNet, including several crucial symptoms as AMN Macular Neuroretinopathy, and Bull's Eye Maculopathy Chloroquine.

\textbf{Contributions.}
\begin{itemize}
    \item We design a novel visual-assisted diagnosis algorithm based on the support vector machine and deep neural networks to facilitate medical diagnosis of retinal diseases. 
    \item We present a new clinical label collection, EyeNet, for Ophthalmology with 32 retina diseases classes. 
    \item Finally, we train a model based on the proposed EyeNet. The consistent diagnostic accuracy of our model would be a crucial aid to the ophthalmologist, and effectively in a point-of-care scenario.
\end{itemize}

\begin{figure}
\begin{center}
   \includegraphics[width=1.0\linewidth]{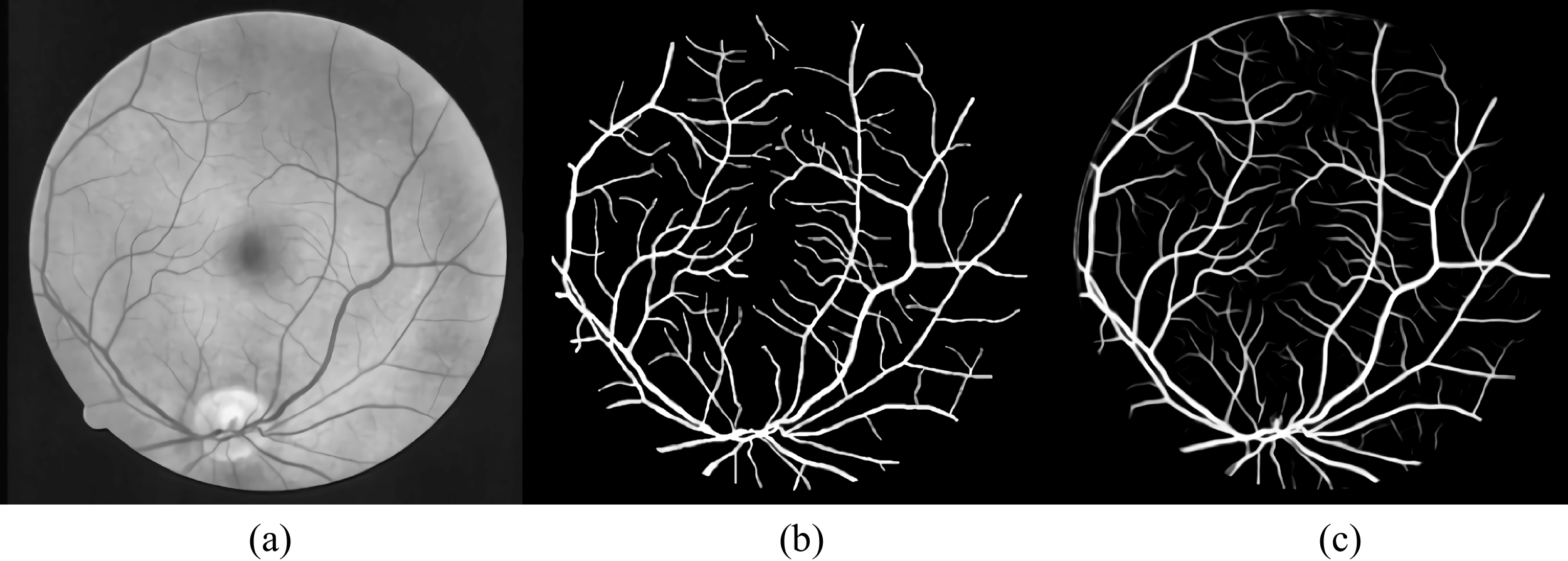}
\end{center}
\vspace{-0.5cm}
   \caption{The figure shows the result of U-Net tested on (a), an unseen eyeball clinical image. (b) is the ground truth and (c) is the generated result of U-Net. Based on (b) and (c), we discover that the generated result is highly similar to the ground truth.}
\label{fig:figure2}
\vspace{-0.3cm}
\end{figure}

\begin{figure}
\begin{center}
   \includegraphics[width=1.0\linewidth]{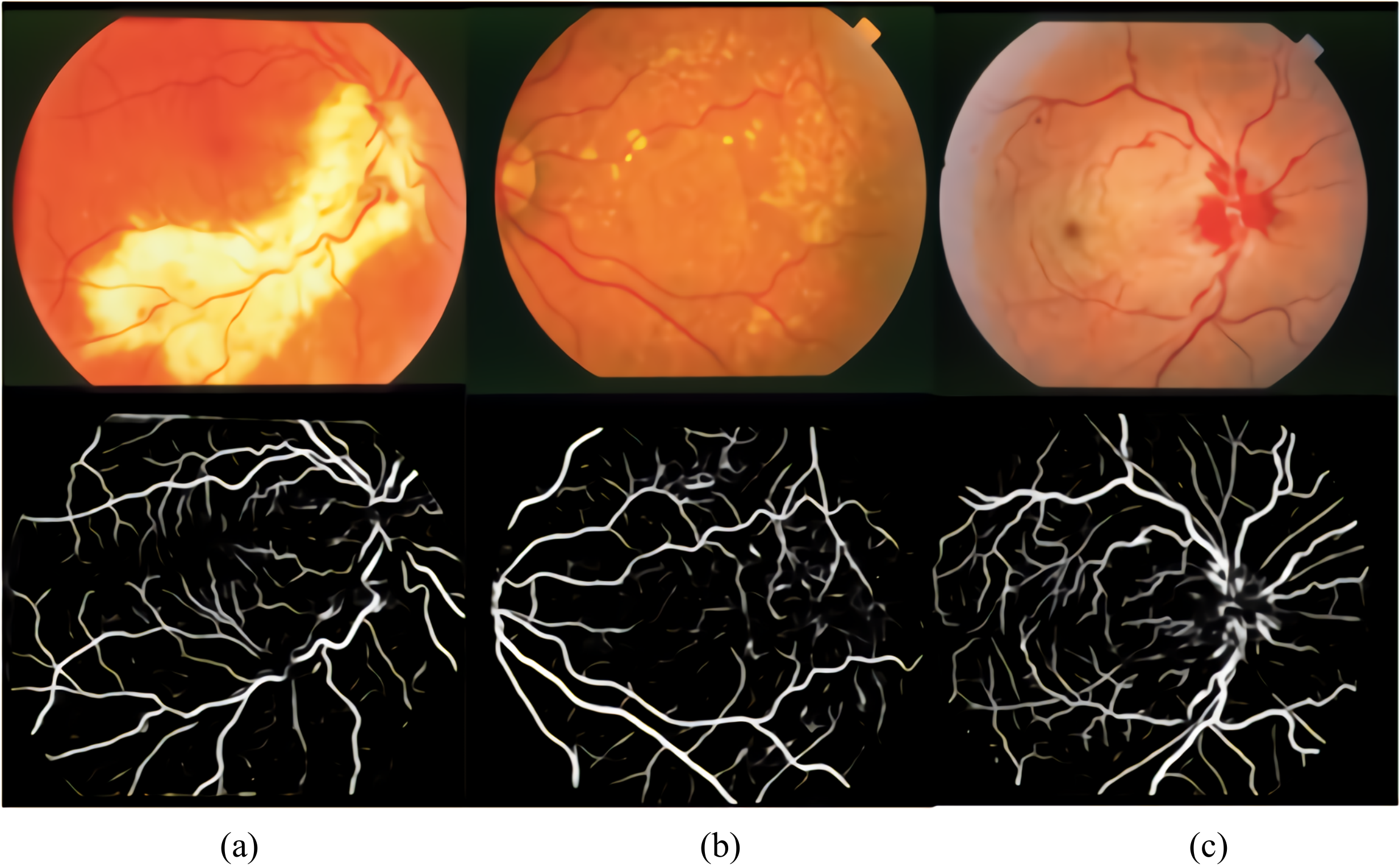}
\end{center}
\vspace{-0.5cm}
   \caption{This figure illustrates the qualitative results of U-Net tested on the colorful images on (a) Myelinated retinal nerve fiber layer (b) age-related macular degeneration (c) disc hemorrhage}
\label{fig:figure3}
\vspace{-0.3cm}
\end{figure}

\vspace{-0.3cm}
\section{Methodology}

In this section, we present the workflow of our proposed model, referring to Figure \ref{fig:figure1}.  

\textbf{2.1.} \textbf{U-Net}

DNNs has greatly boosted the performance of image classification due to its power of image feature learning \cite{simonyan2014very}. Active retinal disease is characterized by exudates around retinal vessels resulting in cuffing of the affected vessels \cite{khurana2007comprehensive}. However, ophthalmology images from clinical microscopy are often overlayed with white sheathing and minor features. Segmentation of retinal images has been investigated as a critical \cite{rezaee2017optimized} visual-aid technique for ophthalmologists. U-Net \cite{ronneberger2015u} is a functional DNNs especially for segmentation. Here, we proposed a modified version of U-Net by reducing the copy and crop processes with a factor of two. The adjustment could speed up the training process and have been verified as an adequate semantic effect on small size images. We use cross-entropy for evaluating the training processes as:
\[
\ E = \sum_{x\in \Omega }w(x)log(p_{l}(x))   \hspace{+1.25cm} (1)
\]

where $p_{l}$ is the approximated maximum function, and the weight map is then computed as:
\[
\ w(x)=w_{c}(x)+w_{0}\cdot exp(\frac{-(d_{x1}+d_{x2})}{2\sigma^2})^2 \hspace{+0.5cm} (2)
\]

 $d_{x1}$ designates the distance to the border of the nearest edges and $d_{x2}$ designates the distance to the border of the second nearest edges. LB score is shown as \cite{cochocki1993neural}. We use the deep convolutional neural network (CNNs) of two $3\times3$ convolutions. Each step followed by a rectified linear unit (ReLU) and a $2\times2$ max pooling operation with stride 2 for downsampling; a layer with an even x- and y-size is selected for each operation. Our proposed model converges at the 44th epoch when the error rate of the model is lower than $0.001$. The accuracy of our U-Net model is 95.27\% by validated on a 20\% test set among EyeNet shown as Figure~\ref{fig:figure2}. This model is robust and feasible for different retinal symptoms as illustrated in Figure~\ref{fig:figure3}.

\textbf{2.2.} \textbf{Principal Component Analysis}

Principal component analysis (PCA) is a statistically matrix-based method by orthogonal transformations. We use PCA combined with SVM classifier to lower the computing complexity and avoid the result of over-fitting on the decision boundary. We optimize SVM classifier with PCA at the $62nd$ principle component.

\textbf{2.3.} \textbf{Support Vector Machine}

Support Vector Machine is a machine learning technique for classification, regression, and other learning tasks. Support vector classification (SVC) in SVM, map data from an input space to a high-dimensional feature space, in which an optimal separating hyperplane that maximizes the boundary margin between the two classes is established.
The hinge loss function is shown as:
 \vspace{-0.1cm}
\[
\ \frac{1}{n}\left [ \sum_{i=1}^{n} max(0,1-y_{i}(\vec{w}\cdot\vec{x_{i}} ))\right ]+\lambda \left \| \vec{w} \right \|^2 \hspace{+0.5cm} (3)
\]
 \vspace{-0.1cm}
Where the parameter $\lambda$ determines the trade off between increasing the margin-size and ensuring that the $\vec{x_{i}}$ lie on the right side of the margin. Parameters are critical for the training time and performance on machine learning algorithms. We pick up cost function parameter c as 128 and gamma as 0.0078. The SVM has comparably high performance when the cost coefficient higher than 132. We use radial basis function (RBF) and polynomial kernel for SVC.

\section{Efforts on Retinal Label Collection}




Retina Image Bank (RIB) is an international clinical project launched by American Society of Retina Specialists in 2012, which allows ophthalmologists around the world to share the existing clinical cases online for medicine-educational proposes. Here we present EyeNet which is mainly based on the RIB. To this end, we manually collected the 32 symptoms from RIB, especially on the retina-related diseases. Different from the traditional retina dataset \cite{staal2004ridge} focused on the morphology analysis, our given open source dataset labeled from the RIB project is concentrated on the difference between disease for feasible aid-diagnosis and medical applications. With the recent success on collecting high-quality datasets, such as ImageNet \cite{krizhevsky2012imagenet}, we believe that collecting and mining RIB for a more developer-friendly data pipeline is valuable for both Ophthalmology and Computer Vision community enabling development of advanced analytical techniques.


\begin{figure}
  \centering
    \includegraphics[width=0.50\textwidth]{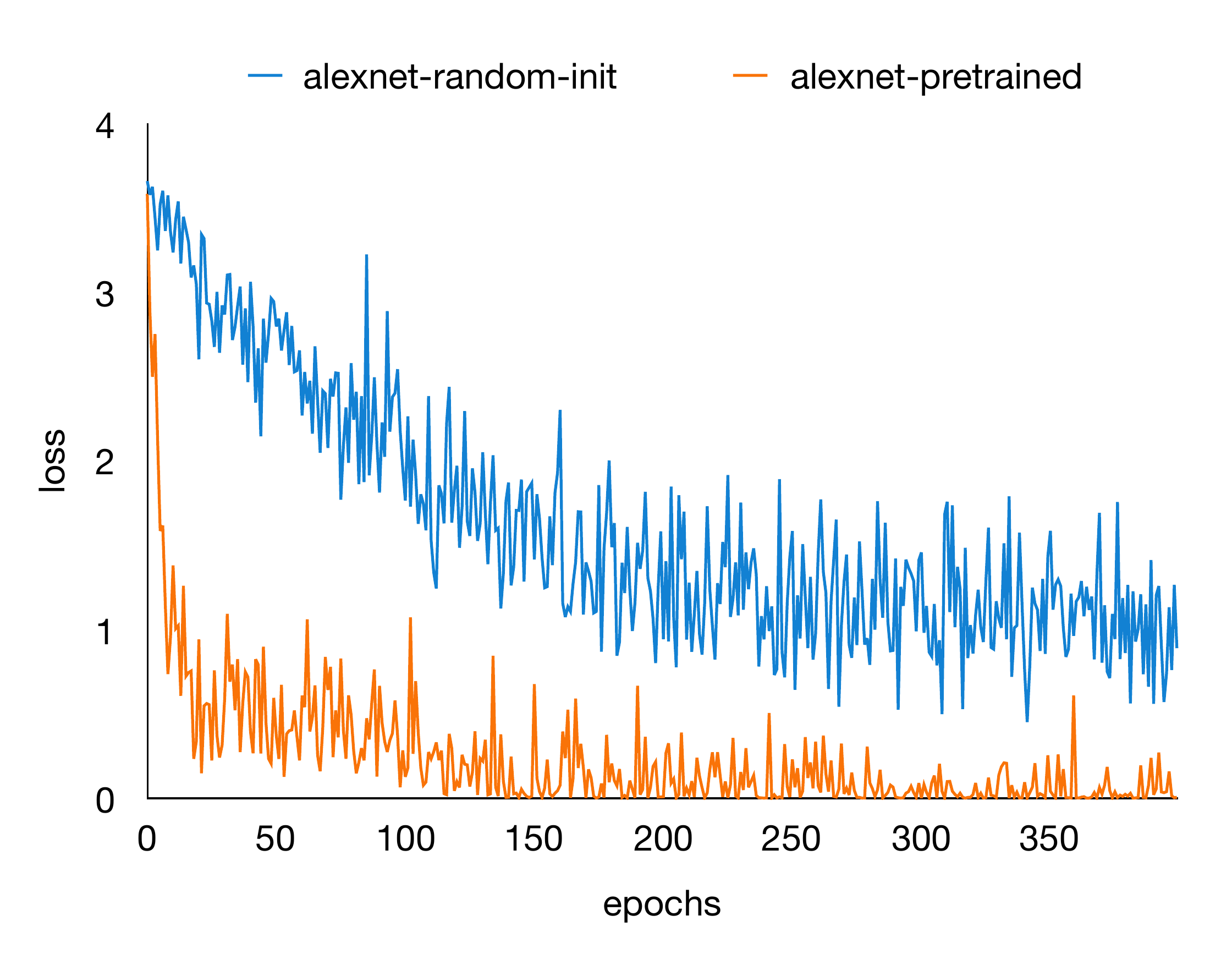}
    \vspace{-1.0cm}
    \caption{A plot of loss training AlexNet initialized with ImageNet pretrained weights (orange) and initialized with random weights (blue).}
\label{fig:figure4}
\vspace{-0.4cm}
\end{figure}

\section{Experiments}

In this section, we describe the implementation details and experiments we conducted to validate our proposed method.

\textbf{4.1.} \textbf{Dataset}

For experiments, the original EyeNet is randomly divided into three parts: 70\% for training, 10\% for validation and 20\% for testing. All the training data have to go through the PCA before SVM. All classification experiments are trained and tested on the same dataset.

\textbf{4.2.} \textbf{Setup}

The EyeNet has been processed to U-Net to generate a subset with the semantic feature of a blood vessel. For the DNNs and Transfer Learning models, we directly use the RGB images from retinal label collection. EyeNet is published online: github.com/huckiyang/EyeNet

\textbf{4.3.} \textbf{Deep Convolutional Neural Networks}

CNNs have demonstrated extraordinary performance in visual recognition tasks \cite{krizhevsky2012imagenet}, and the state of the art is in a great many vision-related benchmarks and challenges \cite{xie2017aggregated}. With little or no prior knowledge and human effort in feature design, it yet provides a general and effective method solving variant vision tasks in variant domains. This new development in computer vision has also shown great potential helping/replacing human judgment in vision problems like medical imaging \cite{esteva2017dermatologist}, which is the topic we try to address in this paper. In this section, we introduce several baselines in multi-class image recognition and compare their results on the EyeNet.

\textbf{4.3.1.} \textbf{Baseline1-AlexNet}

AlexNet \cite{krizhevsky2012imagenet} verified the feasibility of applying deep neural networks on large scale image recognition problems, with the help of GPU. It brought up a succinct network architecture, with 5 convolutional layers and 3 fully-connected layers, adopting ReLU~\cite{nair2010rectified} as the activation function.

\textbf{4.3.2.} \textbf{Baseline2-VGG11}

VGG\cite{simonyan2014very} uses very small filters (3x3) repeatedly to replace the large filters (5x5,7x7) in traditional architectures. By pushing depths of the network, it achieved state-of-the-art results on ImageNet with fewer parameters.

\textbf{4.3.3.} \textbf{Baseline3-SqueezeNet}

Real world medical imaging tasks may require a small yet effective model to adapt to limited resources of hardware. As some very deep neural networks can cost several hundred megabytes to store, SqueezeNet \cite{iandola2016squeezenet} adopting model compression techniques has achieved AlexNet level accuracy with $\sim$500x smaller models.


\textbf{4.4.} \textbf{Transfer Learning}

We exploit a transfer learning framework from normalized ImageNet \cite{krizhevsky2012imagenet} to the EyeNet for solving the small samples issue on the computational retinal visual analytics. With sufficient and utilizable training classified model, Transfer Learning can resolve the challenge of Machine Learning in the limit of a minimal amount of training labels by means of Transfer Learning, which drastically reduce the data requirements. The first few layers of DNNs learn features similar to Gabor filters and color blobs and these features appear not to be specific to any particular task or dataset and thus applicable to other datasets and tasks ~\cite{yosinski2014transferable}. Experiments have shown significant improvement after applying pretrained parameters on our deep learning models, referring to Table \ref{table:table1} and Table \ref{table:table2}.

\textbf{4.5.} \textbf{Hybrid-SVMs Results}

All SVM are implemented in Matlab with libsvm \cite{chang2011libsvm} module. We separate both the original retinal dataset and the subset to three parts included 70\% training set, 20\% test set, and 10\% validation set. By training two multiple-classes SVM models on both original EyeNet and the subset, we implement a weighted voting method to identify the candidate of retina symptom. We have testified different weight ratio as $Hybrid-Ratio$, SVM model with \{RGB Images: SVM model with U-Net subset\}, between EyeNet and the subset with Vessel features to make a higher accuracy at Table \ref{table:table1}. We have verified the model without over-fitting by the validation set via normalization on the accuracy with \~2.31\% difference. 
\begin{table}[t]
\begin{center}
\scalebox{0.975}{
    \begin{tabular}{| l | l | l |}
    \hline
    \textbf{Hybrid-Ratio} & \textbf{RBF kernel}& \textbf{Polyno. kernel} \\ \hline
    ~~~0\%~:~100\%& ~~~~0.8203 & ~~~~~~~0.8439  \\ \hline
    ~~~40\%~:~60\%& ~~~~0.8371 & ~~~~~~~0.8381 \\ \hline
    ~~~47\%~:~53\% & ~~~~\textbf{0.8973} & ~~~~~~~0.8781 \\ \hline
    ~~~61\%~:~39\%& ~~~~0.8903 & ~~~~~~~\textbf{0.8940}  \\ \hline
    ~~~100\%~:~0\%& ~~~~0.8626 & ~~~~~~~0.8733  \\    
    \hline
    \end{tabular}}
    \caption{Accuracy comparison of Hybrid-SVM with RBF and Polynomial kernel. We introduce a hybrid-ratio of the mixed weighted voting between two multi-SVCs trained from EyeNet and the U-Net subset. }
    \vspace{-0.2cm}
\label{table:table1}
\end{center}
\end{table}

\begin{table}[t]
\begin{center}
\scalebox{1.1}{
    \begin{tabular}{| l | l | l |}
    \hline
    ~~~\textbf{Model} & \textbf{Pretrained} & \textbf{Random Init.} \\ \hline
    ~~AlexNet & ~~~~0.7903 & ~~~~~~0.4839  \\ \hline
    ~~VGG11 & ~~~~\textbf{0.8871} & ~~~~~~\textbf{0.7581} \\ \hline
    SqueezeNet & ~~~~0.8226 & ~~~~~~0.5633  \\    
    \hline
    \end{tabular}}
    \caption{Accuracy comparison of three DNNs baselines}
    \label{table:table2}
\vspace{-0.5cm}
\end{center}
\end{table}

\textbf{4.6.} \textbf{Deep Neural Networks Results}

All DNNs are implemented in PyTorch. We use identical hyperparameters for all models. The training lasts 400 epochs. The first 200 epochs take a learning rate of 1e-4 and the second 200 take 1e-5. Besides, we apply random data augmentation during training. In every epoch, there is $70\%$ probability for a training sample to be affinely transformed by one of the operations in \{flip, rotate, transpose\}$\times$\{random crop\}. Though ImageNet and our Retinal label collection are much different, using weights pretrained on ImageNet rather than random ones has boosted test accuracy of any models with 5 to 15 percentages, referring to Table \ref{table:table2}. Besides, pretrained models tend to converge much faster than random initialized ones as suggested in Figure~\ref{fig:figure4}. The performance of DNNs on our retinal dataset can greatly benefit from a knowledge of other domains.

\section{Conclusion and Future Work}

In this work, we have designed a novel hybrid model for visual-assisted diagnosis based on the SVM and U-Net. The performance of this model shows the higher accuracy, 89.73\%, over the other pre-trained DNNs models as an aid for ophthalmologists. Also, we propose the EyeNet to benefit the medical informatics research community. Finally, since our label collection not only contains images but also text information of the images, Visual Question Answering \cite{huang2017vqabq,huang2017novel,huang2017robustness} based on the retinal images is one of the interesting future directions. Our work may also help the remote rural area, where there are no ophthalmologists locally, to screen retinal disease without the help of ophthalmologists in the future.

\section*{Acknowledgement}

This work is supported by competitive research funding from King Abdullah University of Science and Technology (KAUST). Also, we would like to acknowledge Google Cloud Platform and Retina Image Bank, a project from the American Society of Retina Specialists. 



\bibliography{example_paper}
\bibliographystyle{icml2013}

\end{document}